%% file: 0-main.tex
\documentclass[sigconf]{acmart} 
\AtBeginDocument{%
  }

\setcopyright{acmlicensed}
\copyrightyear{2018}
\acmYear{2018}
\acmDOI{XXXXXXX.XXXXXXX}
\acmConference[SETN 2026]{Make sure to enter the correct
  conference title from your rights confirmation email}{September 09--11,
  2026}{Chania, Greece}

\acmISBN{978-1-4503-XXXX-X/2018/06}

\setcopyright{none}
\renewcommand\footnotetextcopyrightpermission[1]{}

\begin{document}

\title[A Review of Counterfactual GNN Explainers]{A Comparative Study of Counterfactual Explainers for Graph Neural Networks Enabling Multiple Types of Graph Edits}


\author{Maria Myrto Villia}
\email{mvillia@ics.forth.gr}
\affiliation{%
  \institution{Institute of Computer Science, Foundation for Research and Technology - Hellas (FORTH)}
  \city{Heraklion}
  \country{Greece}
}
\affiliation{%
  \institution{Computer Science Department, University of Crete}
  \country{Greece}
}

\author{Filippos Gouidis}
\email{gouidis@ics.forth.gr}
\affiliation{%
  \institution{Institute of Computer Science, Foundation for Research and Technology - Hellas (FORTH)}
  \city{Heraklion}
  \country{Greece}
}
\affiliation{%
  \institution{Computer Science Department, University of Crete}
  \country{Greece}
}

\author{Theodore Patkos}
\email{patkos@ics.forth.gr}
\affiliation{%
  \institution{Institute of Computer Science, Foundation for Research and Technology - Hellas (FORTH)}
  \city{Heraklion}
  \country{Greece}
}

\author{Panos Trahanias}
\email{trahania@ics.forth.gr}
\affiliation{%
  \institution{Institute of Computer Science, Foundation for Research and Technology - Hellas (FORTH)}
  \city{Heraklion}
  \country{Greece}
}
\affiliation{%
  \institution{Computer Science Department, University of Crete}
  \country{Greece}
}
\renewcommand{\shortauthors}{Villia et al.}

\begin{abstract}
Counterfactual explanations for graph-structured data seek to determine minimal and realistic modifications required in an input graph to alter a model’s prediction to a predefined output. Although counterfactual explainers that support modifying the graph by both adding and removing edges have recently emerged, there is still a lack of general and efficient methods, especially when considering the quality of the generated explanations. Moreover, the problem remains far from solved, as existing methods exhibit different strengths and weaknesses, often trading off between explanation size, coverage and quality. For this reason, it is important to identify where each method performs well and where it falls short, so as to guide future research in the field. Thus, our study compares six state-of-the-art (SOTA) models on a diverse set of real-world and synthetic datasets, covering both binary and multi-class graph and node classification tasks, and evaluates their performance using diverse quantitative and qualitative metrics.
\end{abstract}




\maketitle

\input{1-Intro}

\input{2-related_work}
\input{3-preliminaries}

\input{4-Experiments}

\input{5-Discussion}

\begin{acks}
This study is funded by the research project CARAML implemented in the framework of H.F.R.I. call ``$3^{rd}$ Call for H.F.R.I.’s Research Projects to Support Faculty Members Researchers'' (H.F.R.I. Project Number: 25735).
\end{acks}


\bibliographystyle{ACM-Reference-Format}
\bibliography{0-main}


\appendix

\section{Hyperparameter Selection}

The Appendix reports the training configurations and dataset-specific hyperparameters used for all methods included in our evaluation. Unless otherwise reported, we follow the original implementations and recommended settings of each method. It should be noted that while some explainers reported results on the same datasets used in our review, we re-trained all frameworks, due to differences in our train/validation/test splits compared to the original evaluations.

\subsection{Graph Classification Hyperparameters}

For \textbf{CF$^2$}, we use $\alpha = 0.7$, $\gamma = 0.9$, a binarization threshold of 0.5, and train for 500 epochs
across all datasets. On BA-2motifs we set $\lambda = 20$ and $\eta = 0.02$; on BA-3motifs and BA-4motifs we increase the
sparsity penalty to $\lambda = 100$ (with $\eta = 0.02$) to account for the larger number of motif classes. For the
real-world datasets (BBBP, SST5, Twitter) we use $\lambda = 20$ and a higher learning rate $\eta = 0.05$. 

For \textbf{D4Explainer}, we train the diffusion model for 800 epochs with learning rate $\eta = 10^{-3}$, decay $\gamma =
0.999$, dropout $= 0.001$, 6 diffusion layers, hidden size 64, and instance normalization. The noise schedule uses
$\sigma_{\text{len}} = 10$ steps with $p_{\text{low}} = 0.0$ and $p_{\text{high}} = 0.4$, and the CF loss weight is
$\alpha_{\text{cf}} = 0.5$ with sparsity level 2.5. 

For \textbf{GIST} we train for 50 epochs with a batch size of 16, hidden dimension 16, and 2 attention heads. The
regularization coefficient is set to $\alpha=0.9$. Optimization is performed with Adam using a learning rate of $10^{-3}$
and weight decay $10^{-5}$. 

\textbf{RSGG-CE} is trained for 30 epochs. Edge candidates are drawn using a positive-and-negative edge sampler
with 500 sampling iterations per step, and one GAN is trained per target class. The generator and discriminator are
optimized with SGD ($\eta=10^{-3}$, batch size 4) under binary cross-entropy loss. The embedding dimension is set to 4 for
BA-2Motifs, BA-2Motifs-3Classes, BA-3Motifs, BA-4Motifs, and BBBP, and to 28 for Graph-SST5 and Twitter. 

For \textbf{GCFExplainer} the coverage trade-off is $\alpha=0.5$. For BA-2Motifs, BA-3Motifs and BA-4Motifs, the decision threshold is $\theta=0.05$, teleport probability $0.20$, and neighborhood sampling size $40$; per-class maximum steps are $[4300,5700]$, $[3600,3000,3400]$, and $[2100,2700,2800,2400]$ respectively. BA-2Motifs-3Classes uses the same settings with steps $[5000,3300,5000]$. For BBBP, $\theta=0.05$, teleport $0.15$, sample size $30$, and steps $[14700,46800]$. For Graph-SST5 and Twitter, $\theta=0.15$, teleport $0.60$, sample size $20$, and 50000 steps per class.

\subsection{Node Classification Hyperparameters}
For \textbf{CF-GNNExplainer}, we use learning rate $10^{-2}$, $\beta = 0.5$ (size-loss weight), and 500 optimisation epochs per node. 

For \textbf{CF$^2$}, the per-node
counterfactual search runs for 2,000 epochs with $\alpha = 0.6$, $\gamma = 0.5$ and $\lambda = 500$. The learning rate is $10^{-2}$ for BA-Shapes, Cora and PubMed,
and $5 \times 10^{-2}$ for Tree-Cycles. 

For \textbf{D4Explainer}, the diffusion model is trained for 800 epochs with batch size 32, learning rate $10^{-3}$ (decay 0.999), dropout 0.001, and a noise
schedule with $\sigma_{\text{len}}=10$, $p_{\text{low}}=0.0$, $p_{\text{high}}=0.4$. For Tree-Cycles the$\alpha_{\text{cf}}=0.5$
and the sparsity penalty is $\lambda_s=2.5$ for Tree-Cycles; for BA-Shapes $\alpha_{\text{cf}}=0.7$ and $\lambda_s=5.0$.

\end{document}

%% file: 1-Intro.tex
\section{Introduction}
\label{intro}

Counterfactual explanations have emerged as a powerful paradigm in explainable Artificial Intelligence (AI), seeking to provide insight into the behavior of black-box, data-driven AI models \cite{JiangLRF24,GuoWXCFLSurvey25,KaddourLL25CausalityCFX}. This perspective is especially valuable for graph-structured data, where explainability is inherently difficult due to the complex interactions between nodes, edges, and features \cite{yuan2023taxonomy}. Thus, the demand for trustworthy AI systems has naturally extended to Graph Neural Networks (GNNs).

Graphs are widely used to represent complex relationships in a variety of domains, including chemical molecules, social networks, and traffic systems. To capture complex and challenging graph structures, GNNs have been introduced as a powerful class of models capable of receiving graphs as input, learning representations, and making predictions. GNNs are able to capture both the structure of the graph and the features of individual nodes to make predictions. Typical downstream tasks addressed by GNNs include graph classification, where an entire graph is assigned a label, e.g., whether a molecule represented as a graph is toxic or not; node classification, where each node is labeled individually, e.g., predicting the topic of a paper in a citation network, and link prediction, which aims to infer missing connections between nodes or complete patterns inside the graphs.  

Early approaches to explainability in GNNs focused on local-level factual methods \cite{dai24Survey,yuan2023taxonomy,agarwal2023graphxai}, which, given an already trained model and an input graph, aim to identify the most influential subgraph responsible for a specific prediction. These methods, which dominated the field of GNN explainability for years, typically assess how the prediction changes when certain components of the graph are removed, highlighting the elements that are crucial for the model’s decision.

Counterfactual explanations (CfXs) on graphs have a much shorter history \cite{GuoWXCFLSurvey25,KaddourLL25CausalityCFX,CFSurveyPradoRASG24}. However, they provide a complementary advantage by going a step further than just identifying important graph features. They indicate not only why a prediction was made, but also what should be changed to alter it. Specifically, counterfactual explanation methods aim to identify the minimal modifications to the graph structure or node features required to alter the prediction of the model. In graph classification, these modifications aim to change the predicted label of the entire graph and may involve perturbations anywhere in the graph. In node classification, they are intended to change the label of a target node, typically by modifying its local neighborhood. Initial attempts of counterfactual GNN explainability borrow insights from the progress achieved in factual methods: in a style similar to factual explainers, they apply various techniques for edge and feature masking, limiting the search for counterfactuals to substructures of the input graph. This approach significantly hinders the informativeness of counterfactuals and often results in out-of-distribution explanations \cite{D4Explainer23}. 

To address these issues, a class of novel methods have been proposed recently, introducing more flexible perturbation strategies, including both edge additions and deletions. Although early works on CfX generation with the ability to both add and delete edges adopted domain-specific assumptions, more generic approaches started to emerge over the last few years, bringing insightful ideas from diverse, yet related fields, such as adversarial attacks \cite{ATEX26,RSGGCE24}, diffusion models \cite{D4Explainer23}, even computer vision-inspired spectral style transfer \cite{GIST25}. Notably, due to the rapid advancement in the field, these systems have not been contrasted against each other yet, neither in recent survey articles \cite{GuoWXCFLSurvey25,KaddourLL25CausalityCFX,CFSurveyPradoRASG24} nor in the experimental evaluation of the individual systems. Yet, this new perspective introduces a significant paradigm shift in the context of counterfactual post-hoc GNN explainability.


To this end, in this paper, we provide a comprehensive comparison of state-of-the-art CfX methods for GNNs. In particular, the contributions of this survey are summarized as follows:
\begin{itemize}

     \item We provide a comprehensive and in depth assessment of state-of-the-art methods for post-hoc, model-agnostic counterfactual explainability for GNNs addressing graph and node classification tasks, extending a recent initial evaluation given in~\cite{DRCFGNN26}.
     
    \item Our study spans a diverse set of commonly employed datasets, including both real-world and synthetic benchmarks, covering binary and multiclass classification settings. More importantly, it adopts a multi-faceted perspective, reporting evaluation metrics that assess not only the performance and efficiency of the explainers, but also the quality of explanations generated.

    \item We offer valuable insights into the research challenges ahead, offering the potential to drive future exploration within a field that is still in its early stages, but has seen a sharp rise in attention over the past few years.
\end{itemize}

%% file: 2-related_work.tex
\section{Literature Classification}
\label{sec:relwork}

\subsection{Factual Explainability of GNNs}
The explainability of GNNs is less explored than that of deep neural models for images and text. Recent review articles paint the picture of the current progress  \cite{yuan2023taxonomy,Longa25survey,dai24Survey}. The majority of GNN explainers are factual, instance-level, post-hoc models. GNNExplainer \cite{ying2019gnnexplainer}, one of the most popular models to date, optimizes soft masks on edges and attributes, aiming to maximize the number of those that can be eliminated while preserving original predictions as much as possible. Other explainers also employ similar perturbation techniques, including SubgraphX \cite{yuan2021subgraphx} which performs Monte Carlo tree search to find the most important subgraph, ZORRO \cite{zorro23} which uses fidelity to review the search process, or PGExplainer \cite{LuoCX20} and GraphMask \cite{SchlichtkrullCT21} that train a deep neural network to predict effective edge perturbation masks. The emphasis on generating subgraphs is shared among all these models. GraphLime \cite{HuangYa23Graphlime}, on the other hand, employs a surrogate model, which can assign large weights to features that are important.

Model-level post-hoc explanation, where the explanation is not tailored to a specific input graph, instead aims to identify generic rules that apply to any input, is more challenging. A notable example is XGNN \cite{YuanT20XGNN} which applies a graph generation module to identify patterns in input graphs that lead to specific predictions. Recently, self-explainable GNNs, such as SE-GNN \cite{DaiWang21} and SES \cite{HuangLW24SES}, have been proposed to provide explanations alongside predictions. Overall, the progress in factual GNN explainability is rapid, creating high expectations for future advancements in the field.

\subsection{Counterfactual Explainability of GNNs}
Contrary to factual GNN explainers, counterfactual explainability in graphs is a relatively recent field, as detailed in recent survey articles \cite{GuoWXCFLSurvey25,KaddourLL25CausalityCFX}. Building off of the momentum gained in factual explainability, most approaches target local-level, post-hoc explanations. For instance, CF-GNNExplainer \cite{lucic2022cf} iteratively optimizes a binary perturbation matrix, while RCExplainer \cite{RobustGNNBCX21} learns the shared decision region of the target GNN across multiple input graphs and generates edge masks trying to ensure that the input graph and the counterfactual graph reside on the opposite side of the decision boundary. MOO \cite{LCL21MooExplainer} and $CF^2$ \cite{TGF22CF2} balance mask learning between factual and counterfactual explanation generation using heuristically determined search algorithms. CFExplainer \cite{chu2024graph}, on the other hand, optimizes mask perturbations to reduce the likelihood of producing the original prediction. These models adapt effectively the different strategies to the requirements of counterfactual explainability, but as already mentioned, they only consider edge removal, leading to explanations that are subgraphs of the input graph. 

Edge addition has been incorporated into LEGIT \cite{BN23LegitExplainer} and MEG \cite{NB21MEGExplainer}, highlighting the benefits of generating CfXs. However, the specification of all possible states needed by their Reinforcement Learning module requires extensive domain knowledge, making it difficult to generalize the approach. CLEAR \cite{MGM22ClearExplainer} adopts a different approach, based on a variational autoencoder architecture, whose regularization term enforces the model to make minimal changes to the graph structure. From a similar perspective, GREASE \cite{CSW25Grease} adopts a Relational Graph Convolutional Architecture to learn a surrogate model to generate CfXs. Moreover, GREASE is domain-specific, particularly designed for recommendation tasks, while CLEAR requires the existence of an underlying causal model of the data, which often is not available. Finally, C2Explainer \cite{c2explainer25} employs domain-independent edge masking while limiting admissible additions to a predefined supergraph.

Recently, the goal of generating CfXs that apply to large proportions of input graphs has been adopted by global counterfactual explainers, such as GCFExplainer \cite{HKM23GCFExplainer} and GlobalGCE \cite{HZZ2025globalGCE}. Despite the different focus, we again observe the same pattern: the former model both adds and removes edges, but adopts a random walks approach, while the latter, aiming to reduce complexity, implements an autoencoder, in order to generate important counterfactual graphs, yet concentrates on subgraph explanations only. 

Building on the above developments, various counterfactual explainers that allow both edge addition and deletion, enabling more general graph transformations and improved performance, have been proposed in recent years. RSGG-CE \cite{RSGGCE24} is a model in which, during training, the generator learns to transform graphs from the explainee class into counterfactual graphs of the opposite class, while the discriminator distinguishes these generated graphs from real graphs belonging to the target class. RSGG-CE manages to overcome the high complexity with an optimized sampling mechanism. More recently, ATEX-CF \cite{ATEX26} also leverages adversarial attack strategies to jointly model edge additions and deletions to generate counterfactual explanations, focusing on node classification tasks. D4Explainer \cite{D4Explainer23} employs a denoising diffusion model to model the underlying distribution of explanation graphs. The method demonstrates strong performance while supporting both factual and counterfactual explanations. However, its high computational complexity and restriction to diffusion over discrete features limit its applicability. Finally, GIST \cite{GIST25} proposes a backtracking mechanism to cross the oracle’s decision boundary, based on spectral style transfer. Although the approach improves performance in binary classification, it generates large explanations and fails to control predictions in multi-label settings.

Notably, most state-of-the-art models, such as D4Explainer, RSGG-CE and GIST, have been evaluated primarily against baselines that only allow edge removal. In our recent study \cite{DRCFGNN26}, DR-CFGNN was introduced, adopting insights from the link prediction research, but focused exclusively on graph classification tasks. The current review offers an extended and more in depth comparative evaluation of recent frameworks for both graph and node classification tasks.

%% file: 3-preliminaries.tex
\section{Preliminaries}
\label{sec:preliminaries}

\begin{figure}[t]
    \centering
    \includegraphics[width=0.5\textwidth]{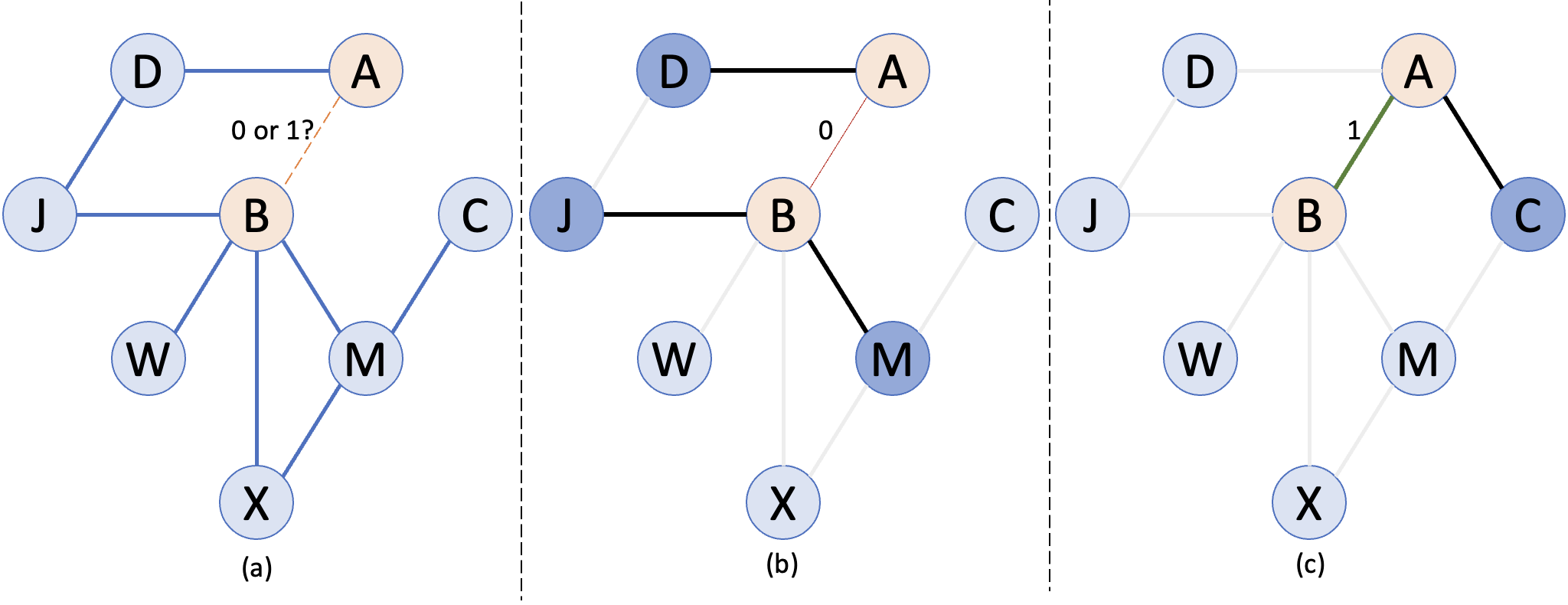}
    \caption{(a) Predictions on drug-drug interaction networks constitute a prominent application domain of Graph Neural Networks (b) subgraph-based explanations identify important edges of the input data (c) We argue that supergraph-based explanations can also offer useful insights}
    \label{Fig:DDInetwork}
\end{figure}

To illustrate the limitations of counterfactual GNN explainers, which are only capable to remove edges, and the important new perspective that the new generation of counterfactual GNN explainers bring, we describe a use case from the domain of bioinformatics. Consider a classifier trained by means of a drug-to-drug interaction network to predict whether an adverse reaction would exist between two drugs if taken together. Such networks are largely incomplete, since examining all chemical reactions by performing clinical or laboratory experiments is impractical due to the potential harms to patients, in addition to being labor-intensive and requiring high financial resources. Recent advances in AI help predict missing edges (correlations), by studying for instance the molecular structure and chemical features of the ingredients. Let us assume that for a particular input graph (Fig. \ref{Fig:DDInetwork}a), the classifier predicts that a combined medication scheme with drugs A and B would be safe; a typical subgraph-based explainer will output those nodes that influence the most the given prediction (Fig. \ref{Fig:DDInetwork}b), and a CfX would aim to modify this subgraph, in order to change the prediction. Nevertheless, an explanation in the form of an expansion of the original graph data may reveal that if an interaction between drug A and drug C existed, the classifier would generate an increased risk of hypersensitivity reaction instead (Fig. \ref{Fig:DDInetwork}c). Such an explanation may constitute a reason for further investigation if a patient is already under treatment with drug C and no definite medical data exist to eliminate such a possibility.

In the sequel, let $G=\{ V, E\} = \{\mathbf{A}, \mathbf{X} \}$ two alternative representations of a graph $G$, where $V = \{v_1, .., v_N\}$ the set of nodes, $E\subseteq V \times V$ the set of edges, $\mathbf{A} \in \{0,1\}^{N \times N}$ the binary adjacency matrix, and $ \mathbf{X} \in \mathbf{R}^{N\times d}$ the feature matrix. Without loss of generality, we define below factual and counterfactual explanation for graph and node classification tasks, which are the most commonly explored tasks among most of the available explainers.

\paragraph{Graph classification :} Let $\Phi:\mathcal{G}\rightarrow \mathcal{Y}$ denote a trained GNN classifier (oracle), where $\mathcal{G}$ is set of graphs and $\mathcal{Y} = {1,\dots,C}$ is the label space. Given a prediction $\Phi(G)= y$, where a single label is assigned to each graph, a factual GNN explainer $\Psi_{\Phi,F}(\cdot)$ seeks to identify one or more subgraphs $G^F \subset G$ that explain the prediction of $\Phi$, such that the predicted label remains unchanged, i.e., $\Phi(G^F) = \Phi(G)$. In contrast, a counterfactual explainer $\Psi_{\Phi,CF}(\cdot)$ seeks to identify one or more counterfactual graphs $G^{CF}$ that differ minimally from $G$ while remaining plausible, and yield a different prediction, i.e., $\Phi(G^{CF}) \neq \Phi(G)$.

According to \cite{PVL24}, a counterfactual for $G$ is within the distribution of valid counterfactuals given by
\begin{equation}
   \mathop{arg max}_{G^{CF} \in \mathcal{G}^{CF}} P(G^{CF} | G, \Phi(G), \neg \Phi(G)  ) 
\end{equation}
where $\mathcal{G^{CF}}$ is the set of all possible counterfactuals by perturbing $G$, and $\neg \Phi(G)$ indicates any other class from $\Phi(G)$. The problem is typically reformulated as an optimization problem that aims to identify minimal graph edits that change the prediction; yet, an acceptable CfX should exhibit satisfactory performance in other aspects too, including plausibility and robustness.

\paragraph{Node classification :} Let $\Phi:\mathcal{G}\times \mathcal{V}\rightarrow \mathcal{Y}$ be a trained GNN classifier (oracle) that, given a graph $G=(\mathcal{V},\mathcal{E})$ and a node $v \in \mathcal{V}$, assigns a label in $\mathcal{Y} = {1,\dots,C}$ to $v$. Given a prediction $\Phi(G, v)=y$, a factual GNN explainer $\Psi_{\Phi,F}(\cdot)$ seeks to identify a subgraph $G^F \subseteq G$ relevant to node $v$ that preserves the prediction, i.e., $\Phi(G^F, v)=\Phi(G, v)$. Similarly, a counterfactual explainer $\Psi_{\Phi,CF}(\cdot)$ aims to identify a subgraph $G^{CF} \subseteq G$ relevant to a node $v \in \mathcal{V}$ that differs minimally from $G$ while remaining plausible, and changes the prediction for $v$, i.e., $\Phi(G^{CF}, v) \neq \Phi(G, v)$.


%% file: 4-Experiments.tex
\section{Evaluation}

\subsection{Counterfactual Explainers}
For \textbf{graph classification} experiments, our study considers four state-of-the-art counterfactual explainers: \textbf{CF$^{2}$} \cite{TGF22CF2}, which generates explanations by only removing edges; \textbf{D4Explainer} \cite{D4Explainer23}, which leverages diffusion models; \textbf{RSGG-CE} \cite{RSGGCE24}, which is based on generative adversarial networks (GANs); and \textbf{GIST} \cite{GIST25}, which relies on style transfer. 
In addition, we include \textbf{GCFExplainer} \cite{HKM23GCFExplainer} as a reference. Unlike the other systems, GCFExplainer is a global-level explainer whose objective is to identify high-level rules that generalize across a large proportion of the input graphs. We also implement a naive exhaustive-search \textbf{Random} baseline that randomly adds and removes edges. For each graph, we explore all combinations of up to two edge deletions and up to two edge additions until a counterfactual example is found. Although this approach is computationally expensive and does not scale to larger graphs, it provides a useful reference point against which the performance of approximate methods can be compared, despite often producing implausible counterfactual explanations (CfXs).

For \textbf{node classification} experiments, we compare \textbf{D4Explainer} and \textbf{CF$^{2}$}. We also use \textbf{CF-GNNExplainer} \cite{lucic2022cf} as a baseline, which is designed specifically for node classification and generates counterfactual explanations by removing edges only. A Random baseline is not considered, as the search space is restricted to the local neighborhood of the target node. Under this setting, random edge perturbations are substantially more likely to affect the prediction, especially when the target node belongs to the ground-truth motif, making such a baseline less informative than in the graph classification setting.

Since not all systems  support both prediction tasks, the explainers are selected separately for graph and node classification according to their intended scope and available implementations. Notably, CF$^{2}$ and D4Explainer are evaluated in both settings, enabling a direct comparison of their behavior across the two prediction tasks. Training configurations and dataset-specific hyperparameter selection used for all methods are given in the Appendix.

\subsection{Datasets}
\label{dataset}
For graph classification, we used $4$ synthetic and $3$ real-world datasets. The \textbf{BA-2Motifs} \cite{LuoCX20} dataset contains $1000$ graphs. Each graph is formed by attaching a five-node cycle or a house motif to a Barabási–Albert (BA) graph. The motif determines the graph label ($0$ for the cycle and $1$ for the house motif). We extend this to \textbf{BA-3Motifs} and \textbf{BA-4Motifs}: the former introduces a five-node clique (label $2$), while the latter additionally includes a four-node clique (label $2$) and a five-node clique (label $3$), alongside the original two classes. \textbf{BA-2Motifs-3classes} is introduced to evaluate the ability of models to handle incomplete motifs, by removing a single edge from the motif in one-third of the BA-2Motifs graphs and assigning them a third label ($2$). All node features are 10-dimensional vectors with constant value $0.1$. \textbf{Graph-SST5} and \textbf{Graph-Twitter} \cite{yuan2023taxonomy} are sentiment analysis datasets with $5$ and $3$ classes, respectively, reflecting the sentiment of the sentence. Each sentence is represented as a graph with word nodes and relational edges. \textbf{BBBP} \cite{martins2012bayesian} is a molecular graph classification dataset for the prediction of blood-brain barrier permeability. Molecules are represented as graphs with atoms as nodes and bonds as edges, $9$-dimensional node features, and binary labels indicating whether a molecule can cross the barrier. Table~\ref{tab:dataset_stats} summarizes the statistics of the datasets used in graph classification experiments.
\begin{table}
\centering
\caption{Dataset statistics.}
\label{tab:dataset_stats}
\setlength{\tabcolsep}{3.5pt} 
\renewcommand{\arraystretch}{1.15}
\resizebox{\linewidth}{!}{%
\begin{tabular}{c c c c c c c c}
\toprule
\textbf{Dataset} &
\textbf{\shortstack{\# of\\ classes}} &
\textbf{\shortstack{\# of\\ features}} &
\textbf{\shortstack{Avg. \#\\ of nodes}} &
\textbf{\shortstack{Avg. \#\\ of edges}} &
\textbf{\shortstack{\# of train\\ graphs}} &
\textbf{\shortstack{\# of val.\\ graphs}} &
\textbf{\shortstack{\# of test\\ graphs}} \\
\midrule
BA-2Motifs            & 2 & 10 & 25 & 25,49 & 800 & 100 & 100 \\
BA-2Motifs-3Classes   & 3 & 10 & 25 & 25,23 & 800 & 100 & 100 \\
BA-3Motifs            & 3 & 10 & 25 & 27,06 & 800 & 100 & 100 \\
BA-4Motifs            & 4 & 10 & 25 & 27,09 & 800 & 100 & 100 \\
BBBP                  & 2 & 9 & 24.06  & 25,35 & 1631 & 203 & 205 \\
Twitter               & 3 & 768 & 21.103 & 20,35 & 4,998 & 1,250 & 692 \\
Graph-SST5            & 5 & 768 & 19.849  & 18,66 & 8,544 &  1,101  & 2,210 \\
\bottomrule
\end{tabular}%
}
\end{table}

For node classification, we used $2$ synthetic and $2$ real-world datasets. \textbf{BA-Shapes} consists of a Barabási–Albert (BA) graph with $80$ attached house motifs, where nodes are labeled according to their position in the motif, or $0$ otherwise. \textbf{Tree-Cycles} consists of a tree graph (nodes with label 0) with cycle motifs (nodes with label 1). \textbf{Cora} and \textbf{PubMed} \cite{yang2016revisiting} are citation networks, where nodes represent publications, edges denote citations, and node labels correspond to research topics, with $7$ and $3$ node classes, respectively. Table~\ref{tab:dataset_stats_nc} summarizes the statistics of the node classification datasets.

\begin{table}[t]
\centering
\caption{Statistics of the node classification datasets.}
\label{tab:dataset_stats_nc}
\setlength{\tabcolsep}{4pt}
\renewcommand{\arraystretch}{1.15}
\resizebox{\linewidth}{!}{%
\begin{tabular}{c c c c c c c c}
\toprule
\textbf{Dataset} &
\textbf{\shortstack{\# of\\ classes}} &
\textbf{\shortstack{\# of\\ features}} &
\textbf{\shortstack{\# of\\ nodes}} &
\textbf{\shortstack{\# of\\ edges}} &
\textbf{\shortstack{\# of train\\ nodes}} &
\textbf{\shortstack{\# of val.\\ nodes}} &
\textbf{\shortstack{\# of test\\ nodes}} \\
\midrule
BA-Shapes   & 4 & 10   & 700   & 2055  & 560  & 70   & 70 \\
Tree-Cycles & 2 & 10   & 871   & 971   & 696  & 87   & 88 \\
Cora        & 7 & 1433 & 2708  & 5278  & 1647  & 479  & 479 \\
PubMed      & 3 & 500  & 19717 & 44324 & 13243 & 3000 & 3000 \\
\bottomrule
\end{tabular}%
}
\end{table}

\subsection{Evaluation Metrics}
\label{evaluations_metrics}

In the following, we present the evaluation metrics used to compare the state-of-the-art models. All experiments were run on a single workstation with an Intel Core i9-14900KS CPU (24 physical cores, 32 threads), 62\,GB system RAM, and one NVIDIA GeForce RTX 4090 GPU.

\paragraph{Validity} Validity measures the proportion of instances for which the explainer generates at least one counterfactual, noting that some explainers may produce multiple counterfactuals per instance. For graph classification, it corresponds to the fraction of the original input graphs for which a counterfactual graph is found. For node classification, it measures the fraction of nodes in the single input graph for which the explainer, by adding or removing edges to the local neighborhood of the node, changes its original predicted label.

\paragraph{Explanation Size} Explanation Size is the total number of edge edits that generate a counterfactual instance.

\paragraph{Fidelity} Fidelity captures the decrease in the probability of the model for the original prediction after counterfactual modifications are applied. For graph classification, this corresponds to the confidence change for the input graph, while for node classification, it is computed for the target node based on modifications in its local neighborhood.

\paragraph{Motif Proximity}  
Motif Proximity measures the proportion of edge modifications in the counterfactual explanation that touch the ground-truth motif. This metric is applicable to synthetic datasets where ground-truth explanation motifs are known, and for node classification tasks only when the target node belongs to a motif.

\paragraph{Minimality} 
Minimality measures the necessity of a counterfactual explanation. Specifically, the metric considers all subsets of the applied edits (edge additions and removals) and computes the proportion of subsets for which the prediction remains unchanged when applied to the original graph.

In the tables, we report the average values of Explanation Size, Fidelity, Minimality, and Motif proximity across all instances for which a counterfactual was generated. In cases where multiple counterfactuals are produced per instance, the best is selected and considered in the average of Explanation Size, Fidelity, and Motif Proximity. For Minimality, we first average the scores over all counterfactuals per instance, and then average across all instances.

\subsection{Results}

\paragraph{Graph Classification}

\input{table_validity_setn}

Table~\ref{tab:pn_sizes} shows the validity and average fidelity scores of all explainers on both synthetic and real datasets for graph classification tasks, along with the average explanation size for the optimal counterfactual (according to each method) and for the first counterfactual found. For explainers that generate only a single counterfactual, a ‘-’ is reported. Regarding validity, both the global GCFExplainer, which leverages cross-dataset rules, and the Random baseline, based on exhaustive search, achieve near-perfect scores. However, on real-world datasets, the validity of the Random baseline decreases due to the increased complexity of the graphs and the resulting timeouts, while GCFExplainer maintains perfect validity at the expense of substantially larger explanations, suggesting that more extensive graph modifications are required to achieve prediction flips in these datasets. GIST has moderate validity and fidelity, but its explanations are too large to be practical for human interpretation. D4Explainer produces counterfactuals of moderate size and achieves strong validity and fidelity scores, likely because it explores the search space extensively. However, this leads to very long execution times (see Table~\ref{tab:times} later on). RSGG-CE also produces counterfactuals of moderate sizes while maintaining strong validity and fidelity scores in synthetic datasets. However, on more complex, real-world data, both validity and fidelity decrease. CF$^{2}$, on the other hand, consistently generates small explanations, but this often comes at the cost of lower validity and fidelity, particularly in real datasets. This suggests that while it favors minimal changes, these are not always sufficient to flip the prediction.

Overall, the results highlight a trade-off between validity, fidelity, and explanation size. This trade-off also relates to the quality of the suggested changes: CfXs should focus on meaningful modifications within an entity’s local neighborhood, rather than arbitrary or distant edits that may lead to out-of-distribution effects. In other words, instead of systems that produce an answer in every case, it is preferable to develop models that generate explanations in as many cases as possible, while keeping explanation size small, maintaining high confidence (i.e., fidelity), and avoiding unnecessary (e.g., reflected by minimality) or unrealistic changes (e.g., reflected by motif proximity).

\input{table_motif_proximity_setn} 
We evaluated the generated CfXs using the Motif Proximity metric (higher is better) across all explainers and datasets (Table~\ref{tab:motif}). There is a clear limitation in many explainers to suggest edits around the important neighborhood (motif). The random baseline achieves low scores, highlighting that despite its high validity and small explanation size, these metrics alone are insufficient to assess the quality of CfXs. CF$^{2}$ outperforms the other methods on the majority of datasets, while RSGG-CE and GIST exhibit the lowest performance across datasets. Notably, while synthetic datasets, for which a single motif is embedded within an otherwise random graph, should constitute a relatively simple setting, as altering the prediction requires modifying the motif itself, it is revealed that most systems fail to consistently identify and operate within this critical region.

\input{table_minimality_setn}
Table~\ref{tab:minimality} reports the average minimality (higher is better) of the explanations in all datasets. A minimal CfX consists of the smallest set of edge additions and removals required to induce a prediction flip, that is, it does not contain redundant changes. The Random baseline achieves the highest scores. This behavior stems from our implementation, where edges are incrementally modified (added or/and deleted) until a counterfactual is found. RSGG-CE demonstrates the best performance in all state-of-the-art explainers in several datasets. CF$^{2}$ and D4Explainer achieve high minimality scores on real-world datasets such as Twitter and Graph-SST5, suggesting that their strategies are better aligned with real more complex datasets. In contrast, GCFExplainer and GIST tend to produce less minimal counterfactual explanations, likely due to the trade-off between minimality and other desirable properties of explainers, such as validity.

\input{table_times_setn}
Finally, Table~\ref{tab:times} reports the training and inference time of each explainer. RSGG-CE has the lowest training times across all datasets among all explainers. GIST and RSGG-CE compete for the lowest inference times; however, GIST suffers from an increased training cost, particularly on the two larger datasets. D4Explainer has the highest training times in all datasets and explainers. This makes it impractical even for relatively simple graphs, a limitation also acknowledged by its authors. GCFExplainer and CF$^{2}$ have moderate training and inference times, although they remain considerably slower than RSGG-CE. Overall, these results underline the importance of considering runtime alongside performance.

\paragraph{Node Classification}

\input{table_nd_validity}

Table~\ref{tab:pn_sizes_nc} reports the validity, average fidelity, and average explanation size of all evaluated explainers across the considered node classification datasets, as well as the explanation size of the first counterfactual found for explainers that generate multiple counterfactuals. CF-GNNExplainer achieves exceptionally high validity on most datasets while consistently generating compact counterfactual explanations. This performance is primarily attributed to its simple edge-removal strategy, which directly disrupts the structural pattern responsible for the original prediction. Similarly to the graph classification setting, CF$^{2}$ continues to favor compact counterfactual explanations in three out of the four datasets, with BA-Shapes being the only exception. However, this often comes at the expense of validity, particularly on Tree-Cycles and Cora. In BA-Shapes, CF$^{2}$ and, especially, D4Explainer generate large explanations. This behavior is likely due to the increased complexity of BA-Shapes, which contains five node classes, resulting in a larger number of possible edge additions and removals around the ground-truth motif that can alter a node's prediction.

\input{table_nd_motif_proximity}
Table~\ref{tab:motif_nc} reports the Motif Proximity scores for the node classification task. Overall, the explainers achieve only moderate scores, suggesting that they do not consistently focus their modifications around the ground-truth motif. CF$^{2}$ is the only method that achieves a near-perfect score on Tree-Cycles, while also staying relatively close to the highest scores on BA-Shapes. This behavior is consistent with the graph classification results, where CF$^{2}$ also achieved the highest or second-highest Motif Proximity across the synthetic datasets. Notably, Tree-Cycles is the simpler of the two datasets, containing only two node classes, which may explain why identifying and modifying the relevant motif is substantially easier.

\input{table_nd_minimality}
Table~\ref{tab:minimality_nc} reports the minimality scores of the counterfactual explanations generated for node classification. No single method consistently achieves the best performance across all datasets. CF-GNNExplainer obtains the highest minimality on BA-Shapes and PubMed, outperforming the other methods considerably on both datasets. D4Explainer performs exceptionally well on Tree-Cycles, where the explanation size remains very small (Table~\ref{tab:pn_sizes_nc}). In contrast, its performance drops to zero on BA-Shapes, where the generated explanations are considerably larger. This suggests that the larger explanations reported in Table~\ref{tab:pn_sizes_nc} contain more redundant modifications, resulting in low minimality. Finally, CF$^{2}$ generally exhibits very low minimality scores across all node classification datasets in comparison to the graph classification setting.

Table \ref{tab:times_nc} reports training and inference times for the node classification tasks, drawing a similar picture with the corresponding performance for graph classification tasks.

%% file: table_validity_setn.tex
\begin{table}[t]
\caption{Validity (top left, $\uparrow$) / Fidelity (top right, $\uparrow$), Explanation Size (bottom left, $\downarrow$) / Explanation Size of the $1^{st}$ cf (bottom right, $\downarrow$) for graph classification.}\label{tab:pn_sizes}
\centering
\resizebox{\linewidth}{!}{%
\begin{tabular}{p{1.8cm}c|c|c|c|c|c|}
\toprule
 & Random &  GCFEx &  CF$^{2}$ & D4Ex & RSGG-CE & GIST \\
\midrule
BA-2Motifs & \shortstack{ 1/ 0.82\\ 1/ -} & \shortstack{ 1/ 0.83\\ 4.3/ 30.09} & \shortstack{ 0.79/ 0.94\\ 5.06/ -} & \shortstack{ 0.88/ 0.73\\ 36.72/ 50.74} & \shortstack{ 0.96/ 0.78\\ 8.53/ -} & \shortstack{ 0.57/ 0.55\\ 29.64/ -}  \\
\midrule
\shortstack[l]{BA-3Motifs-\\[-0.2em]3Classes} & \shortstack{ 1/ 0.64\\ 1/ -} & \shortstack{ 1/ 0.56\\ 1.44/ 32.48} & \shortstack{ 0.49/ 0.79\\ 1.8/ -} & \shortstack{ 1/ 0.77\\ 1/ 2.12} & \shortstack{ 1/ 0.80\\ 8.37/ -} & \shortstack{ 0.73/ 0.55\\ 29.86/ -} \\
\midrule
BA-3Motifs & \shortstack{ 1/ 0.69\\ 1.27/  -} & \shortstack{ 1/ 0.72\\ 1/ 30.12} & \shortstack{ 0.68/ 0.85\\ 12.91/ -} & \shortstack{ 1/ 0.87\\ 8.39/ 19.47} & \shortstack{ 0.76/ 0.78\\ 8.44/ -} & \shortstack{ 0.7/ 0.59\\ 31.41/ -} \\
\midrule
BA-4Motifs & \shortstack{ 1/ 0.78\\ 1.24/ -} & \shortstack{ 1/ 0.75\\ 1/ 29.66} & \shortstack{ 0.61/ 0.95\\ 6.82/ -} & \shortstack{ 1/ 0.89\\ 5.77/ 22.91} & \shortstack{ 1/ 0.80\\ 4.29/ -} & \shortstack{ 0.79/ 0.75\\ 31.09/ -} \\
\midrule
BBBP & \shortstack{ 0.65/ 0.39\\ 1.57/ -} & \shortstack{ 1/ 0.32\\ 16.82/ 35.01} & \shortstack{ 0.22/ 0.44\\ 2.05/ -} & \shortstack{ 0.90/ 0.85\\ 10.49/ 18.97} & \shortstack{ 0.34/ 0.12\\ 4.67/ -} & \shortstack{ 0.38/ 0.29\\ 27.95/ -} \\
\midrule
Twitter & \shortstack{ 0.44/ 0.19\\ 1.42/ -} & \shortstack{ 1/ 0.21\\ 18.32/ 26.9} & \shortstack{ 0.01/ 0.11\\ 1.6/ -} & \shortstack{ 0.82/ 0.69\\ 8.15/ 12.86} & \shortstack{ 0.38/ 0.12\\ 7.87/ -} & \shortstack{ 0.79/ 0.46\\ 32.38/ -} \\
\midrule
Graph-sst5 & \shortstack{ 0.54/ 0.16\\ 1.36/ -} & \shortstack{ 1/ 0.14\\ 17.23/ 22.73} & \shortstack{ 0.06/ 0.11\\ 2.18/ -} & \shortstack{ 0.86/ 0.56\\ 9.01/ 12.72} & \shortstack{ 0.45/ 0.08\\ 8.74/ -} & \shortstack{ 0.8/ 0.42\\ 25.67/ -} \\
\bottomrule
\end{tabular}%
}
\end{table}

%% file: table_motif_proximity_setn.tex
\begin{table}[t]
\caption{Motif proximity evaluation (average score; higher is better, $\uparrow$) for graph classification.}\label{tab:motif}
\centering
\resizebox{\linewidth}{!}{%
\begin{tabular}{p{3cm}cccccc}
\toprule
 & Random &  GCFExplainer & CF$^{2}$ & D4Explainer & RSGG-CE & GIST \\
\midrule
BA-2Motifs & \shortstack{ \textbf{0.6}} &\shortstack{ 0.45} & \shortstack{ \underline{0.59}} & \shortstack{ 0.44} & \shortstack{ 0.32} & \shortstack{ 0.27}  \\

BA-2Motifs-3Classes & \shortstack{ 0.24} &  \shortstack{ \underline{0.4}} & \shortstack{ \textbf{0.59}} & \shortstack{ 0.39} & \shortstack{ 0.26} & \shortstack{ 0.27}  \\

BA-3Motifs & \shortstack{ \textbf{0.6}} & \shortstack{ \underline{0.53}} & \shortstack{ \textbf{0.6}} & \shortstack{ 0.41} & \shortstack{ 0.27} & \shortstack{ 0.31} \\

BA-4Motifs & \shortstack{ \underline{0.65}}& \shortstack{ 0.57} & \shortstack{ \textbf{0.97}} & \shortstack{ 0.42} & \shortstack{ 0.27} & \shortstack{ 0.3}  \\
\bottomrule
\end{tabular}%
}
\end{table}

%% file: table_minimality_setn.tex
\begin{table}[t]
\caption{Average minimality score per graph (higher is better, $\uparrow$) for graph classification.}\label{tab:minimality}
\centering
\resizebox{\linewidth}{!}{%
\begin{tabular}{p{3cm}cccccc}
\toprule
 & Random &  GCFExplainer & CF$^{2}$ & D4Explainer & RSGG-CE & GIST \\
\midrule
BA-2Motifs &  \textbf{1} &  0.55 &  0.38 &  0.4 &  \underline{0.73} &  0.46  \\
BA-2Motifs-3Classes &  \textbf{1} &  \underline{0.5} &  0.42 &  \underline{0.5} &  0.44 &  0.28 \\
BA-3Motifs &  \textbf{1} &  0.5 &  0.44 &  0.4 &  \underline{0.63} &  0.36  \\
BA-4Motifs &  \textbf{1} &  0.5 &  0.29 &  0.39 &  \underline{0.74} &  0.24  \\
BBBP &  \textbf{1} &  0.67 &  0.89 &  0.73 &  \underline{0.99} &  0.7 \\
Twitter &  \textbf{1} &  0.71 &  \underline{0.99} &  0.97 &  0.91 &  0.88  \\
Graph-sst5 &  \textbf{1} &  0.72 &  \underline{0.96} &  0.94 &  0.91 &  0.79  \\
\bottomrule
\end{tabular}%
}
\vspace{-10pt}
\end{table}

%% file: table_times_setn.tex
\begin{table}[t]
\caption{Training time(s) / Avg Inference time per graph(s) for graph classification.}
\label{tab:times}
\centering
\resizebox{\linewidth}{!}{%
\begin{tabular}{lcccccc}
\toprule
Dataset & Random &  GCFExplainer & CF$^{2}$ & D4Explainer & RSGG-CE & GIST  \\
\midrule
BA-2Motifs & -/0.197 & 588.52/2.943 & 101.829/0.509 & 2341.463/7.899 & \textbf{0.612}/\underline{0.033} & \underline{89.099}/\textbf{0.029} \\
BA-2Motifs-3Class & -/0.164 & 293.99/1.47 & 114.55/0.573 & 2151.111/7.312 & \textbf{1.011}/\textbf{0.009} & \underline{88.747}/\underline{0.018} \\
BA-3Motifs & -/14.15 & 572.86/2.864 & 105.568/0.528 & 2133.333/7.887 & \textbf{1.032}/\underline{0.123} & \underline{85.367}/\textbf{0.019 }\\
BA-4Motifs & -/12.73 & 563.83/2.819 & 104.216/0.521 & 2151.111/8.062 & \textbf{1.433}/\textbf{0.005} & \underline{89.033}/\underline{0.019} \\
BBBP & -/43.36 & 4529.8/11.381 & 617.822/1.552 & 19360/6.862 & \textbf{1.286}/\underline{0.591} & \underline{208.886}/\textbf{0.108} \\
Twitter & -/59.42 & 2379.08/1.714 & \underline{641.166}/\underline{0.462} & 47088.393/8.641 & \textbf{8.765}/0.556 & 10059.242/\textbf{0.163}\\
Graph-SST5 & -/46.48 & 3853.75/1.625 & \underline{3393.769}/1.431 & 110568.979/7.146 & \textbf{14.453}/\textbf{0.133} & 24098.251/\underline{0.245} \\
\bottomrule
\end{tabular}
}

\end{table}

%% file: table_nd_validity.tex
\begin{table}[t]
\caption{Validity ($\uparrow$) / Fidelity ($\uparrow$) (top) and GED ($\downarrow$) / Min.\ CF Size ($\downarrow$) (bottom) for node classification.}\label{tab:pn_sizes_nc}
\centering
\resizebox{\linewidth}{!}{%
\begin{tabular}{p{3cm}ccc}
\toprule
 & CF-GNNExplainer & CF$^{2}$ & D4Explainer \\
\midrule
BA-Shapes
  & \shortstack{ \textbf{1.00}/ 0.493\\ \textbf{1.10}/ \textbf{1.03}}
  & \shortstack{ \underline{0.686}/ \underline{0.792}\\ \underline{22.85}/ \underline{20.22}}
  & \shortstack{ 0.530/ \textbf{1.000}\\ 34.9/ 34.9} \\
\midrule
Tree-Cycles
  & \shortstack{ \underline{0.387}/ \textbf{0.984}\\ \textbf{1.00}/ \textbf{1.00}}
  & \shortstack{ 0.352/ \underline{0.954}\\ 4.65/ \textbf{1.00}}
  & \shortstack{ \textbf{1.000}/ 0.198\\ \underline{1.33}/ \underline{1.33}} \\
\midrule
Cora
  & \shortstack{ \textbf{1.00}/ \textbf{0.137}\\ \underline{2.73}/ \underline{2.08}}
  & \shortstack{ \underline{0.042}/ \underline{0.015}\\ \textbf{1.20}/ \textbf{1.00}}
  & \shortstack{ -/ -\\ -/ -} \\
\midrule
PubMed
  & \shortstack{ \textbf{1.00}/ \textbf{0.471}\\ \textbf{2.26}/ \textbf{2.12}}
  & \shortstack{ \underline{0.881}/ -\\ \underline{6.61}/ \underline{6.00}}
  & \shortstack{ -/ -\\ - / -} \\
\bottomrule
\end{tabular}%
}

\end{table}

%% file: table_nd_motif_proximity.tex
\begin{table}[t]
\caption{Motif Proximity for node classification (higher is better, $\uparrow$).}\label{tab:motif_nc}
\centering
\resizebox{\linewidth}{!}{%
\begin{tabular}{p{3cm}ccc}
\toprule
& CF-GNNExplainer & CF$^{2}$ & D4Explainer \\
\midrule
BA-Shapes &  \underline{0.635} &  0.521 &  \textbf{0.637} \\
Tree-Cycles &  0.589 &  \textbf{0.984} &  \underline{0.723} \\
\bottomrule
\end{tabular}%
}

\end{table}

%% file: table_nd_minimality.tex
\begin{table}[t]
\caption{Minimality ratio for node classification (higher is better, $\uparrow$).}\label{tab:minimality_nc}
\centering
\resizebox{\linewidth}{!}{%
\begin{tabular}{p{3cm}ccc}
\toprule
& CF-GNNExplainer & CF$^{2}$ & D4Explainer \\
\midrule
BA-Shapes&  \textbf{0.513} &  \underline{0.011} &  0.000 \\
Tree-Cycles &  \underline{0.500} &  0.004 &  \textbf{0.904} \\
Cora&  \underline{0.066} &  \textbf{0.098} &  -- \\
PubMed &  \textbf{0.223} &  \underline{0.039} &  -- \\
\bottomrule
\end{tabular}%
}

\end{table}

\begin{table}[t]
\caption{Training times / Avg.\ inference time per node(s) for node classification.}
\label{tab:times_nc}
\centering
\resizebox{\linewidth}{!}{%
\begin{tabular}{lccc}
\toprule
Dataset & CF-GNNExplainer & CF$^{2}$ & D4Explainer \\
\midrule
BA-Shapes & 93.87\,/\,2.35 & 506.28\,/\,7.70 & 1932.58\,/\,13.48 \\
Tree-Cycles & 76.66\,/\,2.47 & 435.30\,/\,5.26 & 4440.70\,/\,8.07 \\
Cora & 434.63\,/\,1.04 & 5855.64\,/\,14.25 & --\,/\,-- \\
PubMed & 2868.73\,/\,1.05 & 2189.60\,/\,0.73 & --\,/\,-- \\
\bottomrule
\end{tabular}
}

\end{table}

%% file: 5-Discussion.tex
\section{Open Challenges}

In this paper, we conducted a comprehensive analysis of a new generation of counterfactual, post-hoc GNN explainers, capable of both adding and removing edges, in the attempt to generate valuable explanations. In doing so, we  identify a number of open research challenges that can drive future research.

For start, it is evident that going beyond just edge removal is a crucial ability, especially for the graph classification tasks. Yet, the multi-faceted nature of the problem renders a universally best-performing approach difficult to achieve. To highlight this aspect, the comparative analysis elucidates the importance of exploring a variety of qualitative criteria. RSGG-CE, for instance, stands out for its capacity to generate compact CfXs with high validity, especially considering its time efficiency, but presents moderate performance in all qualitative metrics. D4Explainer, on the other hand, seems to owe its impressive performance on real data to the computationally very intensive diffusion process, which furthermore seems to hurt performance in smaller, simpler graphs. The GCFExplainer, with its design to offer a global view, achieves its goal for high coverage, but at the same time its poor qualitative scores make clear why a deeper view is crucial for CfXs. We expect in the near future to witness a growing interest in this field, probably bringing insights from diverse areas, in order to achieve a holistically acceptable performance.

It is important  to underline the lack of standardized benchmarks and the effort being made in the field to systematically organize the evaluation process. Most works we considered only include some of the most popular explainers as baselines in their comparative analyses, which are not necessarily the most powerful ones. They also apply evaluation metrics that vary significantly in scope. The quality and robustness of explanations are important aspects that, as our assessment showed, should not be overshadowed by good scores in validity and fidelity metrics. From a similar standpoint, the authors in \cite{CFSurveyPradoRASG24} even suggest the organization of public competitions for counterfactual explainers, as is happening for example in the field of data mining, in order to encourage uniform and well-formatted benchmarks.

We also notice a characteristic emphasis on exploring graph classification tasks, overlooking the complexities in addressing other tasks that currently GNNs exhibit notable performance. Counterfactual explanation of node classification or link prediction tasks is often not included in the evaluation of many systems, while for certain systems is not even supported by the underlying methodologies. For those frameworks that are applicable to both graph and node classification problems, the findings exhibit relatively consistent behaviours, although we notice a drop in performance in the ability to explain node classifications. The good performance of the CF-GNNExplainer in the less qualitative metrics (Table \ref{tab:pn_sizes_nc}), despite its lack of edge addition exploration, is an indication that proper adaptations to the type of task can offer leverage to the explainers; yet, an in depth look to more detailed metrics is crucial in this case, too.

In this survey we did not consider more complex structures, such as heterogeneous or dynamic graphs. This is an emerging research field in graph representation learning, with new and highly impactful GNN prediction models being suggested. We expect that future GNN explainers will need to be broad enough to accommodate a much wider variety of graph data.

\section{Conclusions}
In this paper, we present a comparative study of counterfactual GNN explainers that go beyond edge removal by also allowing edge addition. Our results show that, although recent methods leverage more advanced approaches, such as generative models, and have improved performance across multiple metrics, no single explainer consistently performs best across validity, efficiency, explanation size, minimality, and motif proximity. These findings highlight the need for more general, efficient and qualitatively reliable counterfactual explainers, as well as more standardized benchmarks for evaluating them.